\documentclass[10pt,journal,cspaper,compsoc]{IEEEtran}

\usepackage{amssymb,amsfonts,amsmath,subcaption}
\usepackage{hyperref}
\usepackage{amsopn}
\usepackage{bbm}
\usepackage{color}
\usepackage{graphicx}
\usepackage{caption}
\def\p{\mathbb{P}}
\def\e{\mathbb{E}}

\def\R{\mathbb{R}}
\def\one{\mathbf{1}}

\begin{document}
\title{Graph Matching: Relax at Your Own Risk}
%
%
%
%


\author{Vince Lyzinski, 
        Donniell E. Fishkind, 
        Marcelo Fiori, \\
        Joshua T. Vogelstein, 
        Carey E. Priebe, 
        and Guillermo Sapiro, 
\IEEEcompsocitemizethanks{\IEEEcompsocthanksitem V.L. is with Johns Hopkins University Human Language Technology Center of Excellence.  D.E.F. and C.E.P. are with Johns Hopkins University.  J.T.V. is with Johns Hopkins University and the Child Mind Institute.  G.S. is with Duke University.  M.F. is with Universidad de la Rep\'ublica, Uruguay.  Work partially supported by NSF, NIH, and DoD.\protect
}
}

%
%

%

\newtheorem{theorem}{Theorem}
\newtheorem{lemma}[theorem]{Lemma}

\IEEEcompsoctitleabstractindextext{%
\begin{abstract}
Graph matching---aligning a pair of graphs to minimize their edge disagreements---has received wide-spread attention from both theoretical and applied communities over the past several decades, including combinatorics, computer vision, and connectomics.  Its attention can be partially attributed to its computational difficulty.  Although many heuristics have previously been proposed in the literature to approximately solve graph matching, very few have any theoretical support for their performance.  A common technique is to relax the discrete problem to a continuous problem, therefore enabling practitioners to bring gradient-descent-type algorithms to bear.  We prove that an indefinite relaxation (when solved exactly) almost always discovers the optimal permutation, while a common convex relaxation almost always fails to discover the optimal permutation.  These theoretical results suggest that initializing the indefinite algorithm with the convex optimum might yield improved practical performance.  Indeed, experimental results illuminate and corroborate these theoretical findings, demonstrating that excellent results are achieved in both benchmark and real data problems by amalgamating the two approaches.
\end{abstract}
}

\maketitle

\section{Introduction}


\IEEEPARstart{S}{everal} problems related to the isomorphism and matching of graphs have been an important and enjoyable challenge for the scientific community for a long time, 
with applications in pattern recognition (see, for example, \cite{PA2,PA1}), computer vision (see, for example, \cite{CV1,CV3,CV2}), and machine learning (see, for example, \cite{nips4,nips3}), to name a few.  
Given two graphs, the graph isomorphism problem consists of determining whether these graphs are isomorphic or not, that is, if there exists a bijection between the vertex sets of the graphs which exactly preserves the vertex adjacency. The graph isomorphism problem is very challenging from a computational complexity point of view.  Indeed, its complexity is still unresolved: it is not currently classified as NP-complete or P \cite{GandJ}. The graph isomorphism problem is contained in the (harder) graph matching problem. The graph matching problem consists of finding the exact isomorphism between two graphs if it exists, or, in general, finding the bijection between the vertex sets that minimizes the 
number of adjacency disagreements. Graph matching is a very challenging and well-studied problem in the literature with applications in such diverse fields as pattern recognition, computer vision, neuroscience, etc.\@ (see \cite{ConteReview}).  Although polynomial-time algorithms for solving the graph matching problem are known for certain classes of graphs 
(e.g., trees \cite{torsello2005polynomial,ullman1974design}; planar graphs \cite{hopcroft1974linear}; and graphs with some spectral properties \cite{fiori2014spectral,alex}), there are no known polynomial-time algorithms for solving the general case.  Indeed, in its most general form, the graph matching problem is equivalent to the NP-hard quadratic assignment problem.

Formally, for any two graphs on $n$ vertices with respective $n \times n$ adjacency matrices $A$ and $B$,
the graph matching problem is to minimize $\| A - PBP^T \|_F$ over
all $P \in \varPi$, where $\varPi$ denotes the set of $n \times n$
permutation matrices, and $\| \cdot \|_F$ is the Froebenius matrix norm 
(other graph matching objectives have been proposed in the literature as well, this being a common one).
Note that for any permutation matrix $P$, $\frac{1}{2}\|A - PBP^T\|_F^2=
\frac{1}{2} \|AP-PB\|_F^2 $ counts the number of adjacency
disagreements induced by the vertex bijection corresponding to $P$.

An equivalent formulation of the graph matching problem
is to minimize $-\langle AP , PB \rangle$ over~all
$P \in \varPi$, where $\langle \cdot , \cdot \rangle$ is the
Euclidean inner product, i.e., for all $C,D \in \R^{n \times n}$,
$\langle C,D \rangle := \textup{trace}(C^TD)$. This can be seen
by expanding, for any $P \in \varPi$,
\begin{eqnarray*}
\|A -PBP^T\|^2_F &=& \|AP -PB\|_F^2  \\
& = &
 \|A\|_F^2 + \|B\|_F^2 -2 \langle AP, PB \rangle ,
\end{eqnarray*}
and noting that $\|A\|_F^2$ and $\|B\|_F^2$ are constants for the optimization
problem over $P \in \varPi$.

Let ${\mathcal D}$ denote the set of $n \times n$ doubly stochastic
matrices, i.e., nonnegative matrices with row and column sums each equal to $1$.
We define the {\it convex relaxed graph matching problem} to be  minimizing
$\|AD -DB\|_F^2$ over all $D \in {\mathcal D}$, and we define the
{\it indefinite relaxed graph matching problem} to be minimizing
$-\langle AD , DB \rangle$ over all $D \in {\mathcal D}$. Unlike
the graph matching problem, which is an integer programming problem,
these relaxed graph matching problems are each continuous optimization problems with
a quadratic objective function subject to affine constraints.
Since the quadratic objective $\|AD -DB\|_F^2$ is also convex
in the variables $D$ (it is a composition
of a convex function and a linear function), there is a
polynomial-time algorithm for exactly solving the convex relaxed graph
matching problem (see \cite{gold}). However, $-\langle AD , DB \rangle$ is not
convex (in fact, the Hessian has trace zero and is therefore {\it indefinite}), and nonconvex quadratic programming is (in general) NP-hard.  Nonetheless the indefinite relaxation can be
efficiently approximately solved with Frank-Wolfe (F-W) methodology \cite{FW,FAQ}. 

It is natural to ask how the (possibly different) solutions to these relaxed formulations relate to the solution of the original graph matching problem.
Our main theoretical result, Theorem~\ref{thd}, proves, under mild conditions, that convex relaxed graph matching (which is tractable) almost always yields the wrong matching, and indefinite relaxed graph matching (which is intractable) almost always yields the correct matching. We then illustrate via illuminating simulations that this asymptotic result about the trade-off between tractability and correctness is amply felt even in moderately sized instances.

In light of graph matching complexity results (see for example \cite{alex,hardness,atserias2013sherali}), it is unsurprising that the convex relaxation can fail to recover the true permutation.  In our main theorem, we take this a step further and provide an answer from a probabilistic point of view, showing almost sure failure of the convex relaxation for a very rich and general family of graphs when convexly relaxing the graph matching problem.  This paints a sharp contrast to the (surprising) almost sure correctness of the solution of the indefinite relaxation.  We further illustrate that our theory gives rise to a new state-of-the-art matching strategy.

\subsection{Correlated random Bernoulli graphs}
\label{sec:crbg}
Our theoretical results will be set in the context of correlated random (simple) Bernoulli graphs,\footnote{Also known as \textit{inhomogeneous random graphs} in \cite{boll}.} which can be used to model many real-data scenarios.  Random Bernoulli graphs are the most general edge independent random graphs, and contain many important random graph families including Erd\H os-R\'enyi and the widely used stochastic block model of \cite{sbm} (in the stochastic block model, $\Lambda$ is a block constant matrix, with the number of diagonal blocks representing the number of communities in the network).
Stochastic block models, in particular, have been extensively used to model networks with inherent community structure (see, for example, \cite{snijders1997estimation,nowicki2001estimation,newman2004finding,airoldi2009mixed}).  As this model is a submodel of the random Bernoulli graph model here used, our main theorem (Theorem \ref{thd}) extends to stochastic block models immediately, making it of highly practical relevance.

  These graphs are defined as follows.
Given $n\in\mathbb{Z}^+$, a real number $\rho \in [0,1]$, and a
symmetric, hollow matrix $\Lambda \in [0,1]^{n \times n}$, define $\mathcal{E}~:=~\{ \{i,j\}:i\in[n],j\in[n],i\neq j\},$ where $[n]:=\{1,2,\ldots,n\}.$
Two random graphs with respective $n \times n$
adjacency matrices $A$ and $B$ are
{\it $\rho$-correlated Bernoulli$(\Lambda)$} distributed if, for
all $\{i,j\}\in \mathcal{E}$, the random variables (matrix entries)
$A_{i,j},B_{i,j}$ are Bernoulli$(\Lambda_{i,j})$ distributed,
and all of these random variables are collectively independent except that, for each
 $\{i,j\}\in \mathcal{E}$, the Pearson product-moment
 correlation coefficient for $A_{i,j},B_{i,j}$ is $\rho$.
It is straightforward to show that the parameters $n$, $\rho$, and $\Lambda$
completely specify the random graph pair distribution, and the distribution may be
achieved by first, for all $\{i,j\}\in\mathcal{E}$,
having $B_{ij} \sim \textup{Bernoulli}(\Lambda_{i,j})$ independently drawn and
then, conditioning on $B$, have
$A_{i,j}\sim \textup{Bernoulli}\left (  (1-\rho)\Lambda_{i,j}+ \rho B_{i,j}
 \right )$ independently drawn. While $\rho=1$ would imply the graphs are isomorphic, this model allows for a natural vertex alignment (namely the identity function) for $\rho<1$, i.e. when the graphs are not necessarily isomorphic.

\subsection{The main result}
We will consider a sequence of 
correlated random Bernoulli~graphs for $n=1,2,3,\dots$, where $\Lambda$ 
is a function of $n$. When we say that a sequence of events, $\{E_m\}_{m=1}^{\infty}$, holds {\it almost always} we mean that 
almost surely it happens that the events in the sequence occur for all but finitely many $m$. 

 \begin{theorem} \label{thd}
{\it Suppose $A$ and $B$ are adjacency matrices
for $\rho$-correlated Bernoulli$(\Lambda)$ graphs,
and there is an $\alpha \in (0,1/2)$ such that
$\Lambda_{i,j} \in [\alpha,1-\alpha]$ for all $i\ne j$.
Let $P^* \in \varPi$, and denote $A':=P^* A P^{*T}$.\\
a) If $(1-\alpha)(1-\rho)<1/2$, then it almost always holds that
\[\arg \min_{D \in {\mathcal D}}  -\langle A'D,DB \rangle
 = \arg \min_{P \in \varPi} \|A'-PBP^T\|_F   =  \{ P^* \} .\]
b) If the between graph correlation $\rho <\! 1$, then it almost always holds that 
$P^* \not \in
\arg\min_{D \in {\mathcal D}}  \|A'D-DB\|_F .$}
\end{theorem}

This theorem states that: (part {\it a}) the unique solution of the indefinite relaxation almost always is the correct permutation matrix, while (part {\it b}) the correct permutation is almost always not a solution of the commonly used convex relation. 
Moreover, as we will show in the experiments section, the convex relaxation can lead to a doubly stochastic matrix that is not even in the Voronoi cell of the true permutation.  
In this case, the convex optimum is closest to an incorrect permutation, hence the correct permutation will not be recovered by projecting the doubly stochastic solution back onto $\varPi$.

In the above, $\rho$ and $\alpha$ are fixed.  However, the proofs follow {\it mutatis mutandis} if $\rho$ and $\alpha$ are allowed to vary in $n$.  If there exist constants $c_1,c_2>0$ such that 
$\alpha\geq c_1\sqrt{(\log n)/n}$ and $1/2-c_2\sqrt{(\log n)/n}\geq (1-\rho)(1-\alpha),$ then Theorem \ref{thd}, part {\it a} will hold.  Note that $\alpha\geq c_1\sqrt{(\log n)/n}$ also guarantees the corresponding graphs are almost always connected.
For the analogous result for part {\it b}, let us first define 
$\sigma(i)=\frac{1}{n-1}\sum_{k\neq i} \Lambda_{ki}(1-\Lambda_{ki}).$ 
If there exists an $i\in [n]$ such that 
$ 1-\frac{3}{2\sigma(i)}\sqrt{(8\log n)/n}>\rho,$ then the results of Theorem \ref{thd}, part {\it b} hold as proven below.

\subsection{Isomorphic versus $\rho$-correlated graphs}	
There are numerous algorithms available in the literature for (approximately) solving the graph isomorphism problem (see, for example, \cite{cordella2004sub,fankhauser2012suboptimal}), as well as for (approximately) solving the subgraph isomorphism problem (see, for example, \cite{ullmann1976algorithm}).  All of the graph matching algorithms we explore herein can be used for the graph isomorphism problem as well.

We emphasize that the $\rho$-correlated random graph model extends our random graphs beyond isomorphic graph pairs; indeed $\rho$-correlated graphs $G_1$ and $G_2$ will almost surely have on the order of 
$[\alpha,1-\alpha]\rho n^2$ edge-wise disagreements.  As such, these graphs are a.s. {\it not} isomorphic.  In this setting, the goal of graph matching is to align the vertices across graphs whilst simultaneously preserving the adjacency structure as best possible across graphs.  However, this model does preserve a very important feature of isomorphic graphs:  namely the presence of a latent alignment function (the identity function in the $\rho$-correlated model).

We note here that in the $\rho$-correlated Bernoulli($\Lambda$) model, both $G_1$ and $G_2$ are marginally Bernoulli$(\Lambda)$ random graphs, which is amenable to theoretical analysis. We note here that real data experiments across a large variety of data sets (see Section \ref{data}) and simulated experiments across a variety of robust random graph settings (see Section \ref{sec:otherrand}) also both support the result of Theorem \ref{thd}.  Indeed, we suspect that an analogue of Theorem \ref{thd} holds over a much broader class of random graphs, and we are presently investigating this extension.

\section{Proof of Theorem \ref{thd}, part a} Without loss of generality, let $P^*=I$.  
We will first sketch the main argument of the proof, and then we will spend the remainder of the section filling in all necessary details of the proof.  
The proof will proceed as follows. 
Almost always, $-\langle A,B\rangle<-\langle AQ,PB\rangle$ for any $P,\,Q\in\varPi$ such that either $P\neq I$ or $Q\neq I$.  
To accomplish this, we count the entrywise disagreements between $AQ$ and $PB$ in two steps (of course, this is the same as the number of entrywise disagreements between $A$ and $PBQ^T$).  We first count the entrywise disagreements between $B$ and $PBQ^T$ (Lemma \ref{thy}), and then count the additional disagreements induced by realizing $A$ conditioning on $B$.  Almost always, this two step realization will result in more errors than simply realizing $A$ directly from $B$ without permuting the vertex labels (Lemma \ref{ths}).  This establishes $-\langle A,B\rangle<-\langle AQ,PB\rangle$, and Theorem \ref{thd}, part {\it a} is a consequence of the Birkhoff-von Neumann theorem.

We begin with two lemmas used to prove Theorem \ref{thd}.
First, Lemma \ref{thg} is adapted from \cite{alon}, presented here
as a variation of the form found in
\cite[Prop. 3.2]{kim}.  This lemma lets us tightly estimate the number of disagreements between $B$ and $PBQ^T$, which we do in Lemma \ref{thy}.
\begin{lemma} 
{\it For any integer $N>0$ and constant $\alpha \in (0,\frac{1}{2})$,
suppose that the random variable $X$ is a function of at most $N$
 independent Bernoulli random variables, each with Bernoulli parameter in the interval $[\alpha, 1-\alpha]$.  Suppose that changing the value of any one of the
Bernoulli random variables (and keeping all of the others fixed)
changes the value of $X$ by at most $\gamma$. Then for any
$t$ such that $0 \leq t< \sqrt{\alpha(1-\alpha)}\gamma N$, it holds that
${\mathbb P} \left [ |X- \mathbb{E}X| > t
\right ] \leq 2 \cdot \text{exp}\{-t^2/(\gamma^2 N)\}$. \label{thg}}
\end{lemma}

The next result, Lemma \ref{lem:hoeff1},
is a special case of the classical Hoeffding inequality (see, for example, \cite{fcci}), which we use to tightly bound the number of additional entrywise disagreements between $AQ$ and $PB$ when we realize $A$ conditioning on $B$.  
\begin{lemma} \label{lem:hoeff1}{\it Let $N_1$ and $N_2$ be positive integers,
and 
 $q_1$ and $q_2$ be real numbers in $[0,1]$.
If $X_1 \sim \textup{Binomial}(N_1,q_1)$ and
$X_2 \sim \textup{Binomial}(N_2,q_2)$ are independent, then for any $t \geq 0$ it holds that
\begin{align*}
\mathbb{P} \Big [ \Big | X_1+X_2 &- \mathbb{E} \Big (X_1+X_2
\Big ) \Big | \geq t
 \Big ] 
 \leq 2 \cdot \textup{exp}\left\{\frac{-2t^2}{N_1+N_2}\right\}.
 \end{align*}} 
\end{lemma}

Setting notation for the next lemmas, let $n$ be given.
Let $\varPi$ denote the set~of~$n \times n$~permutation matrices.
Just for now, fix any $P,Q \in \varPi$ such that
they are not both the identify matrix, and let $\tau, \omega$ be
their respective associated permutations on $[n]$; i.e. for all
$i,j \in [n]$ it holds that $\tau(i)=j$ precisely when $P_{i,j}=1$ and,
for all $i,j \in [n]$, it holds that $\omega(i)=j$ precisely when $Q_{i,j}=1$.
It will be useful to define the following sets:
\begin{align*}
\Delta&:= \{ (i,j) \in [n] \times [n] : \tau(i) \ne i \mbox{ or }
\omega(j) \ne j  \},\\
\Delta_t&:= \{ (i,j) \in \Delta : \tau(i)=j
\mbox{ and } \omega(j)=i \},\\
\Delta_d&:= \{ (i,j) \in \Delta : i = j \mbox{ or }
\tau(i)=\omega(j)\},\\
\Delta_\tau&:=\{ (i,j) \in [n] \times [n] : \tau(i) \ne i \},\\
\Delta_\omega&:=\{ (i,j) \in [n] \times [n] :
\omega(j) \ne j  \}.
\end{align*}
If we define $m$ to be the maximum of
$| \{ i \in [n]: \tau(i) \ne i \} |$ and
$| \{ j \in [n]: \omega(j) \ne j \} |$, then it follows that
$mn \leq | \Delta | \leq 2mn$.  
This is clear from noting that $\Delta_{\omega},\Delta_\tau\subseteq\Delta\subseteq\Delta_\tau\cup\Delta_\omega$.
Also, $|\Delta_t|\leq m$, since for $(i,j)\in\Delta_t$ it is necessary that $\tau(i)\neq i$ and $\omega(j)\neq j$.
Lastly, $|\Delta_d|\leq 4m$, since
$$\Delta_d\subseteq\{(i,i)\in\Delta\}\cup \{ (i,j) \in \Delta : i\neq j,\,
\tau(i)=\omega(j)\},$$
and $|\{(i,i)\in\Delta\}|\leq 2m$, and $|\{ (i,j) \in \Delta : i\neq j,\,
\tau(i)=\omega(j)\}|\leq 2m$.

We make the following assumption in all that follows:\\
{\bf Assumption 1:}  {\it Suppose that $\Lambda \in [0,1]^{n \times n}$ is a
symmetric, hollow matrix, there is a real number
$\rho \in [0,1]$, and there is a constant
$\alpha \in (0,1/2)$ such that $\Lambda_{i,j} \in [\alpha,1-\alpha]$
for all $i\ne j$, and $(1-\alpha)(1-\rho)<1/2$.
Further, let $A$, $B$ be the adjacency matrices
of two random $\rho$-correlated Bernoulli$(\Lambda)$ graphs.}

Define the (random) set 
$$\Theta' := \{ (i,j) \in \!\Delta: \!i \!\ne\!j,\text{ and }B_{i,j} \ne B_{\tau(i),\omega(j)} \}.$$  Note that $|\Theta'|$ counts the entrywise disagreements induced {\it within} the off-diagonal part of $B$ by $\tau$ and $\omega.$

\begin{lemma} \label{thy}
{\it Under Assumption 1, if $n$ is sufficiently large then 
$$ \p \left ( | \Theta' | \not \in \left [
\alpha mn/3, \ 2mn
\right  ]  \right ) \leq
 2e^{-\alpha^2mn/128} . $$}
\end{lemma}

\noindent {\bf Proof of Lemma \ref{thy}:}
For any $(i,j) \in \Delta$, note that
$(B_{i,j}-B_{\tau(i),\omega(j)})^2$ has a Bernoulli
distribution; if $(i,j) \in \Delta_t\cup \Delta_d,$
then the Bernoulli parameter is either $0$ or is in the interval
$[\alpha, 1-\alpha]$, and if $(i,j) \in
\Delta \backslash (\Delta_t\cup \Delta_d)$,
then the Bernoulli parameter is
$\Lambda_{i,j}(1-\Lambda_{\tau(i),\omega(j)})+(1-\Lambda_{i,j})
\Lambda_{\tau(i),\omega(j)},$
and this Bernoulli parameter
is in the interval $[\alpha,1-\alpha]$
since it is a convex combination of values in this interval. 
Now, $|\Theta'|= \sum_{(i,j)\in \Delta, i \ne j}
(B_{i,j}-B_{\tau(i),\omega(j)})^2$,
so we obtain that 
$
\alpha \left ( |\Delta|-|\Delta_t|-|\Delta_d| \right )
\leq  \e ( | \Theta' | ) \leq (1-\alpha) |\Delta |,
$
 and thus
\begin{equation} \label{thx}
 \alpha m (n-5)  \ \leq  \  \e(| \Theta' |) \ \leq \ 2(1-\alpha)mn.
\end{equation}

Next we apply Lemma \ref{thg}, since
$|\Theta'|$ is a function of the at-most $N:=2mn$ Bernoulli
random variables $\{ B_{i,j} \}_{(i,j)\in \Delta: i \ne j}$,
which as a set (noting that $B_{i,j}=B_{j,i}$ is counted at most once for each $\{i,j\}$) are independent, each with Bernoulli parameter in
$[\alpha,1-\alpha]$.
Furthermore, changing the value of any one
of these random variable would change $|\Theta'|$ by at most $\gamma:=4$, thus
Lemma \ref{thg} can be applied and,
for the choice of $t:=\frac{\alpha}{2}mn$, we obtain that
\begin{equation} \label{thzz}
\p \left [  \big |  |\Theta'| - \e (|\Theta'|) \big | > \alpha mn/2
\right ] \leq 2e^{-\alpha^2mn/128}.
\end{equation}
Lemma \ref{thy} follows from (\ref{thx}) and (\ref{thzz}),
since
\begin{align*}
&\p \big [  \big |  |\Theta'| - \e (|\Theta'|) \big | > \alpha mn/2
\big ]\\
&=\p \left[  |\Theta'|\notin\left[\e (|\Theta'|)-\alpha mn/2,\e (|\Theta'|)+\alpha mn/2\right]\right]\\
&\geq \p \left [  |\Theta'|\notin\left[\alpha m(n-5)-\alpha mn/2,2(1-\alpha)mn+\alpha mn/2\right]
\right ]\\
&\geq \p \left [  |\Theta'|\notin\left[\alpha m(n-5)-\alpha mn/2,2mn\right]\right],
\end{align*}
and $5\alpha mn /6\leq \alpha m(n-5)$
when $n$ is sufficiently large (e.g.\@ $ n \geq 30).\,\,\blacksquare$ 

With the above bound on the number of (non-diagonal) entrywise disagreements between $B$ and $PBQ^T$, we next count the number of additional disagreements introduced by realizing $A$ conditioning on $B$.
In Lemma \ref{ths}, we prove that this two step realization will almost always result in more entrywise errors than simply realizing $A$ from $B$ without permuting the vertex labels.
\begin{lemma} \label{ths}{\it
Under Assumption 1, it almost always holds that,
for all $P,Q \in \varPi$ 
such that either $P \neq I$ or $Q \neq I$, 
$\|A - PBQ^T \|_F > \| A - B \|_F$.}
\end{lemma}

\noindent {\bf Proof of Lemma \ref{ths}:} 
Just for now, let us
fix any $P,Q \in \varPi$ such that either $P \neq I$ or $Q \neq I$, and say
$\tau$ and $\omega$ are their respective associated
permutations on $[n]$.  Let $\Delta$ and $\Theta'$ be
defined as before.
For every $(i,j) \in \Delta$, a combinatorial argument, combined with $A$ and $B$ being binary valued,
yields (where for an event $C$, $\mathbbm{1}_C$ is the indicator random variable for the event $C$)\vspace{-1.3mm}
\begin{align}
\label{eq:comb}
&\mathbbm{1}_{A_{i,j} \ne B_{i,j}} + \mathbbm{1}_{B_{i,j} \ne B_{\tau(i),
\omega(j)}} = \\
&\hspace{10mm}\mathbbm{1}_{A_{i,j} \ne B_{\tau(i),\omega(j)}} +
2 \cdot \mathbbm{1}_{A_{i,j} \ne B_{i,j} \ \& \ B_{i,j} \ne B_{\tau(i),
\omega(j)}}.\notag
\end{align}
Note that 
\begin{align*}
\|A\!-\!PBQ^T\|_F^2&\!=\!\sum_{i,j}(A_{i,j}\!-\!B_{\tau(i),\omega(j)})^2\!=\!\sum_{i,j}\!\mathbbm{1}_{A_{i,j} \ne B_{\tau(i),\omega(j)}}\\
\|A\!-\!B\|_F^2&\!=\!\sum_{i,j}(A_{i,j}\!-\!B_{i,j})^2\!=\!\sum_{i,j}\!\mathbbm{1}_{A_{i,j} \ne B_{i,j}}.
\end{align*}
Summing Eq. (\ref{eq:comb}) over the relevant indices then yields that
\begin{eqnarray} \label{tha}
\|A-PBQ^T\|_F^2-\|A-B\|_F^2= |\Theta|-2|\Gamma|,
\end{eqnarray}
where the sets $\Theta$ and $\Gamma$ are defined as
\begin{align*}
\Theta &:= \{ (i,j)\in[n]\times[n]: B_{i,j} \ne B_{\tau(i),\omega(j)} \}\subseteq\Delta,\\
\Gamma &:=\{ (i,j) \in \Theta : A_{i,j}\ne B_{i,j} \}.
\end{align*}
Now, partition $\Theta$ into sets $\Theta_1$,
$\Theta_2$, $\Theta_d$, and  partition
$\Gamma$ into sets $\Gamma_1$,
$\Gamma_2$ where 
\begin{align*}
\Theta_1&:= \{ (i,j)\in \Theta : i \ne j \mbox{ and } (j,i) \not \in \Theta \},\\
\Theta_2&:= \{ (i,j)\in \Theta : i \ne j \mbox{ and } (j,i) \in \Theta \},\\
\Theta_d&:= \{ (i,j) \in \Theta : i=j \}, \\
\Gamma_1&:= \{ (i,j) \in \Theta_1: A_{i,j} \ne B_{i,j} \}, \\
\Gamma_2&:= \{ (i,j) \in \Theta_2: A_{i,j} \ne B_{i,j} \}.\end{align*}
Note that all $(i,j)$ such that $i=j$ are not in $\Gamma$.
Also note that $\Theta'\subseteq \Theta$ can be partitioned into the disjoint union $\Theta'=\Theta_1\cup\Theta_2$.

Equation (\ref{tha}) implies 
\begin{align*}
|\Gamma_1| + |\Gamma_2| <  (
|\Theta_1| + &|\Theta_2|  )/2\Rightarrow|\Gamma| <  |\Theta|/2\Rightarrow\\
&\|A-B\|_F^2 < \|A-PBQ^T\|_F^2.
\end{align*}
In particular,
\begin{align}
\label{thz}
  \big \{ \| A &-B \|_F \geq \|A-PBQ^T\|_F \big \}  \Rightarrow \notag\\
 &\left \{ |\Gamma_1|+|\Gamma_2| \geq (|\Theta_1| + |\Theta_2|)/2=|\Theta'|/2
 \right \}.
\end{align}

%
Now, conditioning on $B$ (hence, conditioning on $\Theta'$),
we have, for all $i\ne j$, that (see Section \ref{sec:crbg}),
$A_{i,j}\sim\text{Bernoulli}\left((1-\rho)\Lambda_{i,j}+
\rho B_{i,j}\right).$ 
Thus
${\mathbbm 1}_{A_{i,j} \ne B_{i,j}}$ has a Bernoulli distribution
with parameter bounded above by $(1-\alpha)(1-\rho)$.
Thus, $|\Gamma_1|$ is stochastically dominated by a
Binomial$\left ( |\Theta_1|,(1-\alpha)(1-\rho) \right )$ random variable,
and the independent random variable
$|\Gamma_2|$ is stochastically dominated by a
Binomial$\left (
|\Theta_2|,(1-\alpha)(1-\rho) \right )$ random variable.
An application of Lemma \ref{lem:hoeff1} with $N_1:=|\Theta_1|$,
$N_2:=|\Theta_2|$, $q_1=q_2:=
(1-\alpha)(1-\rho)$, and $t:=\left ( \frac{1}{2}- (1-\alpha)(1-\rho)
\right ) |\Theta'| $, yields (recall that we are conditioning on $B$ here)
\begin{align}
\label{thq}
&\p \left [ |\Gamma_1|+|\Gamma_2| \geq |\Theta'|/2
 \right ]\notag \\
   &=\!
\p \!\left [ \!|\Gamma_1|\!+\!|\Gamma_2|\! - \!(1\!-\!\alpha)(1\!-\!\rho)|\Theta'|\!
\geq\! \Big (\! 1\!/2\! -\!(1\!-\!\alpha)(1\!-\!\rho)  \Big ) |\Theta'|
 \right ]\notag \\
& \leq  2\text{exp}\left\{ \frac{-2 \left ( 1/2-(1-\alpha)(1-\rho) \right )^2
|\Theta'|^2 }{|\Theta_1|+ |\Theta_2|  }\right\}\notag\\
&\leq 
2\text{exp}\left\{-2\Big(1/2-(1-\alpha)(1-\rho)\Big)^2 |\Theta'|\right\}.
\end{align}
No longer conditioning
(broadly) on $B$, Lemma \ref{thy}, equations (\ref{thz})
and (\ref{thq}), and $(1-\alpha)(1-\rho)<\frac{1}{2}$,
imply that
\begin{align} 
&\p  \Big [ \| A -PBQ^T \|_F \leq \|A-B\|_F \Big ] \notag \\
& \leq  \p \left ( | \Theta' | \not \in  \big [
\alpha mn/3, \ 2mn
\big ]  \right ) \notag\\
&\hspace{15mm}
+\p  \Big [ |\Gamma_1|+|\Gamma_2| \geq \frac{1}{2}|\Theta'| \
\Big | \ | \Theta' |  \in \big [
\frac{\alpha}{3} mn, \ 2mn
\big ]  \Big ]  \notag \\
&\leq4\,\text{exp}\left\{- \min \bigg\{ \frac{\alpha^2}{128},\frac{2\alpha}{3}
\bigg(\frac{1}{2}-(1-\alpha)(1-\rho)\bigg)^2\bigg\}    mn\right\}.     \label{thb}
\end{align}

Until this point, $P$ and $Q$---and their associated permutations
$\tau$ and $\omega$---have been fixed.
Now, for each $m\in[n]$, define ${\mathcal H}_m$ to be
the event that $\| A-PBQ^T\|_F \leq \|A-B\|_F$
for {\it any} $P,Q \in \varPi$ with the property that their associated
permutations $\tau,\omega$ are such that the maximum
of $| \{ i \in [n]: \tau(i)\ne i \} |$  and
$| \{ j \in [n]: \omega(j)\ne j \} |$ is exactly $m$.
There are at most ${n \choose m}m!{n \choose m}m! \leq n^{2m}$ such
permutation pairs.  

By (\ref{thb}), for every $m\in[n]$, setting 
$$c_1=\min \{ \alpha^2/128,2\alpha
(1/2-(1-\alpha)(1-\rho))^2/3\},$$
we have
$\p ( {\mathcal H}_m )\leq n^{2m} \cdot 4\,
\text{exp}\left\{- c_1    mn\right\}\leq \text{exp}\{-c_2n\},$
for some positive constant $c_2$ (the last inequality holding when $n$ is
large enough). Thus, for sufficiently large $n$, $\p ( \cup_{m=1}^n {\mathcal H}_m ) \leq n\cdot\text{exp}\{-c_2n\}$
decays exponentially in $n$, and is thus finitely summable over
$n=1,2,3,\ldots$.  Lemma \ref{ths} follows from the Borel-Cantelli
Lemma. $\blacksquare$

\noindent {\bf Proof of Theorem \ref{thd}, part a:}
By Lemma \ref{ths}, it almost always follows that for
every $P,Q \in \varPi$ not both the identity, 
$ \langle AQ,PB \rangle < \langle A,B \rangle $.
By the Birkhoff-von Neuman Theorem, ${\mathcal D}$ is the convex hull
of $\varPi$, i.e., for every $D \in {\mathcal D}$, there exists
constants $\{ a_{D,P} \}_{P \in \varPi}$ such that
$D=\sum_{P \in \varPi}a_{D,P}P$ and $\sum_{P \in \varPi}a_{D,P}=1$.
Thus, if $D$ is not the identity matrix, then almost always
\begin{eqnarray*}
\langle AD, DB \rangle & = & \sum_{P \in \varPi} \sum_{Q \in \varPi} a_{D,P}a_{D,Q}
 \langle AQ,PB  \rangle \\
& <& \sum_{P \in \varPi} \sum_{Q \in \varPi} a_{D,P}a_{D,Q}
 \langle A,B  \rangle=\langle A,B \rangle ,
\end{eqnarray*}
and almost always $\text{argmin}_{D\in\mathcal{D}}-\langle AD,DB\rangle=\{I\}$.  $\blacksquare$

\section{Proof of Theorem \ref{thd}, part b}
The proof will proceed as follows:  we will use Lemma \ref{lem:hoeff2} to prove that the identity is almost always not a KKT (Karush-Kuhn-Tucker) point of the relaxed graph matching problem.  Since the relaxed graph matching problem is a constrained optimization problem with convex feasible region and affine constraints, this is sufficient for the proof of Theorem \ref{thd}, part b.

First, we state Lemma \ref{lem:hoeff2}, a variant of Hoeffding's inequality, which we use to prove Theorem \ref{thd}, part b.
\begin{lemma} \label{lem:hoeff2} {\it Let $N$ be a positive integer.  Suppose that
the random variable $X$ is the sum of $N$ independent random
variables, each with mean $0$ and each taking values in the
real interval $[-1,1]$.  Then for any $t \geq 0$, it holds
that $$\mathbb{P}[ |X| \geq t ]\leq 2 \cdot e^{\frac{-t^2}{2N}}.$$}
\end{lemma}
Again, without loss of generality, we may assume $P^*=I$.
We first note that the convex relaxed graph matching problem can be written as
\begin{align}
\min \,&\| AD-DB\|_F^2\label{convexequation},\\
\text{s.t. } &D \one=\one\label{equalconst1},\\
&\one^T D=\one^T\label{equalconst2},\\
&D\geq 0\label{inequalconst},
\end{align}
where (\ref{convexequation}) is a convex
function (of $D$) subject to affine constraints
(\ref{equalconst1})-(\ref{inequalconst}) (i.e., $D \in {\mathcal D}$).  It follows that if $I$ is the global (or local) optimizer of the convex relaxed graph matching problem, then $I$ must be a KKT (Karush-Kuhn-Tucker) point (see, for example, 
\cite[Chapter 4]{baz}).

The gradient of $\|AD-DB\|_F^2$ (as a function of $D$) is
$$\boldsymbol\nabla(D):=2(A^TAD+DBB^T-A^TDB-ADB^T).$$
Hence, a $\widehat D$ satisfying (\ref{equalconst1})-(\ref{inequalconst}) (i.e., $\widehat D$ is primal feasible) is a KKT point if it satisfies
\begin{equation}
\label{KKTcondgeneral}
\boldsymbol\nabla(\widehat D)+\boldsymbol\mu+\boldsymbol\mu'-\boldsymbol\nu=0,
\end{equation}
where $\boldsymbol \mu,$ $\boldsymbol \mu',$ and $\boldsymbol \nu$ are as follows:
\[\boldsymbol\mu:=\left [
\begin{array}{cccc}
\mu_1 & \mu_1 & \cdots & \mu_1 \\
\mu_2 & \mu_2 & \cdots & \mu_2 \\
   \vdots    &  \vdots      &   \ddots     & \vdots       \\
\mu_n & \mu_n & \cdots & \mu_n
\end{array}
\right ]\in\mathbb{R}^{n\times n},\]
noting that the dual variables $\mu_1, \mu_2, \ldots, \mu_n$ are
not restricted.  They correspond to the equality primal constraints (\ref{equalconst1})
that the row-sums of a primal feasible $D$ are all one;
\[
\boldsymbol\mu':=\left [
\begin{array}{cccc}
\mu'_1 & \mu'_2 & \cdots & \mu'_n \\
\mu'_1 & \mu'_2 & \cdots & \mu'_n \\
   \vdots    &  \vdots      &   \ddots     & \vdots       \\
\mu'_1 & \mu'_2 & \cdots & \mu'_n
\end{array}
\right ]\in\mathbb{R}^{n\times n},\]
noting that 
the dual variables $\mu'_1, \mu'_2, \ldots, \mu'_n$ are
not restricted.  They correspond to the equality primal constraints (\ref{equalconst2})
that the column-sums of a primal feasible $D$ are all one;
\[
\boldsymbol\nu:=\left [
\begin{array}{cccc}
0 & \nu_{1,2} & \cdots & \nu_{1,n} \\
\nu_{2,1} & 0 & \cdots & \nu_{2,n} \\
   \vdots    &  \vdots      &  \ddots      & \vdots       \\
\nu_{n,1} & \nu_{n,2} & \cdots & 0
\end{array}
\right ]\in\mathbb{R}^{n\times n},
\]
noting that the dual
variables $\nu_{i,j}$ are restricted to be nonnegative.  They
correspond to the inequality primal constraints (\ref{inequalconst}) that the entries of a primal feasible $D$
be nonnegative.  Complementary slackness further constrains the $\nu_{i,j},$ requiring that $\widehat D_{i,j}\nu_{i,j}=0$ for all $i,j.$

At the identity matrix $I$, the gradient $\boldsymbol\nabla(I)$, denoted $\boldsymbol\nabla$, simplifies
to 
$\boldsymbol\nabla =[\nabla_{i,j}]= 2A^2+2B^2-4AB\in\R^{n\times n};$ 
and $I$ being a KKT point is equivalent to: 
\begin{equation}
\label{KKTcond}
\boldsymbol\nabla+\boldsymbol\mu+\boldsymbol\mu'-\boldsymbol\nu=0,\end{equation}
where $\boldsymbol\mu,\,\boldsymbol\mu',\, \text{ and }\boldsymbol\nu$ are as specified above.
At the identity matrix, complimentary slackness translates to having
$\nu_{1,1}=\nu_{2,2}=\cdots=\nu_{n,n}=0$.

Now, for Equation (\ref{KKTcond}) to hold,
it is necessary that there exist $\mu_1,\mu_2,
\mu'_1,\mu'_2$ such that
\vspace{-3mm}
\begin{eqnarray}
\nabla_{1,1} + \mu_1 + \mu'_1 & = & 0 ,\label{tif} \\
\nabla_{2,2} + \mu_2 + \mu'_2 & = & 0 ,\label{tig} \\
\nabla_{1,2} + \mu_1 + \mu'_2 & \geq & 0 ,\label{tih} \\
\nabla_{2,1} + \mu_2 + \mu'_1 & \geq & 0 \label{tii} .
\end{eqnarray}
Adding equations (\ref{tih}), (\ref{tii}) and subtracting 
equations (\ref{tif}), (\ref{tig}), we obtain\vspace{-3mm}
\begin{eqnarray} \label{thj}
\nabla_{1,2}+\nabla_{2,1} \geq \nabla_{1,1} + \nabla_{2,2}.
\end{eqnarray}
Note that $\frac{1}{2} \boldsymbol\nabla + \frac{1}{2} \boldsymbol\nabla^T =
2(A-B)^T(A-B)$, hence Equation (\ref{thj}) is equivalent to (where $X:=(A-B)^T(A-B)$)
\begin{align} \label{tik}
&2[X]_{1,2}\geq 
[X]_{1,1}+[X]_{2,2}.
\end{align}
Next, referring back to the joint distribution of $A$ and $B$ (see Section \ref{sec:crbg}),
we have, for all $i \ne j$,
\begin{align*}
\mathbb{P}\big [A_{i,j}=0,\,B_{i,j}=1 \big ] &=
\mathbb{P}\big [A_{i,j}=1 ,\,B_{i,j}=0 \big ]
 \\
&=(1-\rho)\Lambda_{i,j}(1-\Lambda_{i,j}).
\end{align*}
Now, since 
\begin{align*}
[X]_{1,1}+[X]_{2,2}=\sum_{i\neq 1} (A_{i,1}-B_{i,1})^2+\sum_{i\neq 2} (A_{i,2}-B_{i,2})^2,
\end{align*}
is the sum of $(n-1)+(n-1)$ Bernoulli random variables which are collectively
independent---besides the two of them which are equal, namely $(A_{12}-B_{12})^2$ and $(A_{21}-B_{21})^2$---we have that
$[X]_{1,1}+[X]_{2,2}$
is stochastically greater than or equal to
a $\text{Binomial}\big (2n-3,2(1-\rho)\alpha(1-\alpha) \big )$
random variable.  Also note that 
$$[X]_{1,2}=\sum_{i\neq 1,2}(A_{i,1}-B_{i,1})(A_{i,2}-B_{i,2})$$
is the sum of $n-2$ independent random variables (namely, the $(A_{i,1}-B_{i,1})(A_{i,2}-B_{i,2})$'s) each with mean $0$ and
each taking on values in $\{-1,0,1\}$. 
Applying Lemma \ref{lem:hoeff1}
and Lemma \ref{lem:hoeff2}, respectively, to
$ X_{11}+X_{22}$ and to
$X_{12}$, with
$t:=(2n-3)2(1-\rho)\alpha(1-\alpha)/4$,
yields
\begin{align*}
&\mathbb{P} \big(
2[X]_{1,2} \geq 
[X]_{1,1}+ [X ]_{2,2}
\big)\\
&\hspace{10mm}\leq 
\mathbb{P} \big(  2[X]_{1,2} \geq 2t  \big) +
\mathbb{P} \big(
[X ]_{1,1}+ [X]_{2,2}
\leq 2t \big)\\
& \hspace{10mm}\leq  2 \cdot e^{\frac{-2t^2}{2n-3}}+ 2 \cdot e^{\frac{-t^2}{2(n-2)}}
\leq e^{-cn},
\end{align*}
for some positive constant $c$ (the last inequality holds when $n$ is large
enough). Hence the probability that Equation (\ref{tik}) holds is seen to
decay exponentially in $n$, and is finitely summable over $n=1,2,3,\ldots$.  Therefore, by the Borel-Cantelli Lemma we have that almost always
Equation~(\ref{tik}) does not hold. Theorem \ref{thd}, part {\it b} is now shown,
since Equation (\ref{tik}) is a necessary condition for
$I \in \arg \min_{D \in {\mathcal D}}  \| AD-DB \|_F^2$.  $\blacksquare$
\section{Experimental results} 

In the preceding section, we presented a theoretical result exploring the trade-off between tractability and correctness when relaxing the graph matching problem.  
On one hand, we have an optimistic result (Theorem~\ref{thd}, part {\it a}) about an indefinite relaxation of the graph matching problem. 
However, since the objective function is nonconvex, there is no efficient algorithm known to exactly solve this relaxation. 
On the other hand, Theorem \ref{thd}, part {\it b}, is a pessimistic result about a commonly used efficiently solvable convex relaxation, which almost always provides an incorrect/non-permutation solution.

After solving (approximately or exactly) the relaxed problem, the solution is commonly projected to the nearest permutation matrix.  
We have not theoretically addressed this projection step yet.  It might be that, even though the solution in $\mathcal{D}$ is not the correct permutation, it is very close to it, and the projection step fixes this.  We will numerically illustrate this not being the case.

We next present simulations that corroborate and illuminate the presented theoretical results, address the projection step, and provide intuition and practical considerations for solving the graph matching problem.
Our simulated graphs have $n=150$ vertices and follow the Bernoulli model described above, where the entries of the matrix $\Lambda$ are i.i.d. uniformly distributed in $[\alpha,1-\alpha]$ with $\alpha = 0.1$.  In each simulation, we run $100$ Monte Carlo replicates for each value of $\rho$.  Note that given this $\alpha$ value, the threshold $\rho$ in order to fulfill the hypothesis of the first part of Theorem \ref{thd} (namely that $(1-\alpha)(1-\rho)<1/2)$ is $\rho = 0.44$.  As in Theorem \ref{thd}, for a fixed $P^*\in\varPi$, we let $A':=P^* AP^{*T}$, so that the correct vertex alignment between $A'$ and $B$ is provided by the permutation matrix $P^*$. 

We then highlight the applicability of our theory and simulations in a series of real data examples.  In the first set of experiments, we match three pairs of graphs with known latent alignment functions.  We then explore the applicability of our theory in matching graphs without a pre-specified latent alignment.  Specifically, we match 16 benchmark problems (those used in \cite{FAQ,Zaslavskiy2009}) from the QAPLIB library of \cite{qaplib}.  See Section \ref{data} for more detail.  As expected by the theory, in all of our examples a smartly initialized local minimum of the indefinite relaxation achieves best performance.

\begin{table}[t!]
\caption{Notation} 
\centering 
\begin{tabular}{l | l| l } 
\hline 
{\bf Notation} & {\bf Algorithm used} & {\bf Ref.}  \\ [0.5ex] 
\hline 
$D^*\in\text{argmin}_{D \in {\mathcal D}}  \| A'D-DB \|_F^2$ & F-W algorithm & \cite{FW},  \\
 & run to convergence &\cite{Zaslavskiy2009}\\ 
 \hline
$P_c=$ projecting $D^*$ to $\Pi$ & Hungarian algorithm & \cite{hungarian}  \\
\hline
FAQ:$P^*$ & FAQ init. at $P^*$ & \cite{FAQ}  \\
\hline
FAQ:$D^*$ & FAQ init. at $D^*$ & \cite{FAQ}  \\
\hline
FAQ:$J$ & FAQ init. at $J$ & \cite{FAQ}  \\
\hline 
\end{tabular}
\label{table:not} 
\end{table}

We summarize the notation we employ in Table \ref{table:not}.  To find $D^*$, we employ the F-W algorithm (\cite{FW, Zaslavskiy2009}), run to convergence, to exactly solve the convex relaxation.  We also use the Hungarian algorithm (\cite{hungarian}) to compute $P_c$, the projection of $D^*$ to $\varPi$.  To find a local minimum of $\min_{D \in {\mathcal D}}-\langle A'D,DB \rangle$, we use the FAQ algorithm of \cite{FAQ}. We use FAQ:$P^*$, FAQ:$D^*$, and FAQ:$J$ to denote the FAQ algorithm initialized at $P^*$, $D^*$, and $J:=\one \cdot\one^T/n$ (the barycenter of $\mathcal{D}$).  We compare our results to the GLAG and PATH algorithms, implemented with off-the-shelf code provided by the algorithms' authors.  
We restrict our focus to these algorithms (indeed, there are a {\it multitude} of graph matching algorithms present in the literature) as these are the prominent relaxation algorithms; i.e., they all first relax the graph matching problem, solve the relaxation, and then project the solution onto $\Pi$.

\subsection{On the convex relaxed graph matching problem}
\label{sec:conv}

Theorem \ref{thd}, part {\it b}, states that we cannot, in general, expect $D^*=P^*$. However, $D^*$ is often projected onto $\Pi$, which could potentially recover $P^*$.  
Unfortunately, this projection step suffers from the same problems as rounding steps in many integer programming solvers, namely that the distance from the best interior solution to the best feasible solution is not well understood.

\begin{figure}[t!]
\centering
\def\svgwidth{270pt}
\begin{scriptsize}
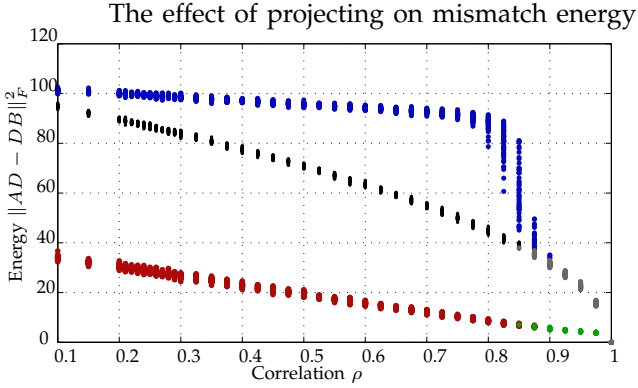
\end{scriptsize}
\caption{For $\rho\in[0.1,1]$, we plot $\|A'D^*-D^*B\|_F^2$~(red /green) and $\|A'P_c-P_cB\|_F^2$ (blue/gray).  Red/blue dots correspond to simulations where $P_c\neq P^*$, and grey/green dots to $P_c=P^*$.  Black dots correspond to $\| A'P^*-P^*B \|_F^2$.  For each $\rho,$ we ran $100$ MC~replicates.}
\label{fig:energy}
\end{figure}

In Figure \ref{fig:energy}, we plot $\| A'D^*-D^*B \|_F^2$ versus the correlation between the random graphs, with $100$ replicates per value of $
\rho$. Each experiment produces a pair of dots, either a red/blue pair or a green/grey pair. 
The energy levels corresponding to the red/green dots correspond to $\| A'D^*-D^*B \|_F^2$, while the energies corresponding to the blue/grey dots correspond $\|A'P_c-P_cB\|_F^2$.  
The colors indicate whether $P_c$ was (green/grey pair) or was not (red/blue pair) $P^*$. 
The black dots correspond to the values of $\| A'P^*-P^*B \|_F^2$.

Note that, for correlations $\rho<1$, $D^*\neq P^*$, as expected from Theorem \ref{thd}, part {\it b}.  Also note that, even for correlations greater than $\rho=0.44$, we note $P_c\neq P^*$ after projecting to the closest permutation matrix, even though with high probability $P^*$ is the solution to the unrelaxed problem. 

We note the large gap between the pre/post projection energy levels when the algorithm fails/succeeds in recovering $P^*$, 
the fast decay in this energy (around $\rho\approx 0.8$ in Figure \ref{fig:energy}), and the fact that the value for $\| A'P^*-P^*B \|_F^2$ can be easily predicted from the correlation value.  These together suggest that $\| A'P_c-P_cB \|_F^2-\| A'D^*-D^*B \|_F^2$ can be used \textit{a posteriori} to assess whether or not graph matching recovered $P^*$.  This is especially true if $\rho$ is known or can be estimated.

How far is $D^*$ from $P^*$?  When the graphs are isomorphic (i.e., $\rho=1$ in our setting), then for a large class of graphs, with certain spectral constraints, then $P^*$ is the unique solution of the convex relaxed graph matching problem \cite{alex}. Indeed, in Figure \ref{fig:energy}, when $\rho=1$ we see that $P^*=D^*$ as expected.  
On the other hand, we know from Theorem \ref{thd}, part {\it b} that if $\rho<1,$ it is often the case that $D^*\neq P^*$.  We may think that, via a continuity argument, if the correlation $\rho$ is very close to one, then $D^*$ will be very close to $P^*$, and $P_c$ will probably recover $P^*$.

\begin{figure}[t]
\centering
\def\svgwidth{270pt}
\begin{scriptsize}
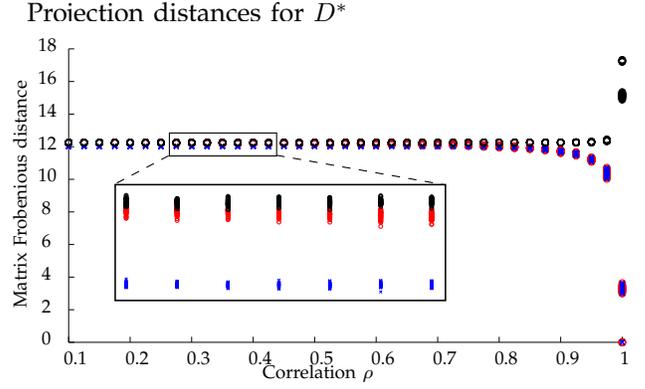
\end{scriptsize}
\caption{Distance from $D^*$ to $P_c$ (in blue), to $P^*$ (in red), and to a random permutation (in black).  For each value of $\rho$, we ran 100 MC replicates.}
\label{fig:dist}
\end{figure}

We empirically explore this phenomena in Figure \ref{fig:dist}.  For $\rho\in[0.1,1]$, with 100 MC replicates for each $\rho$, we plot the (Frobenius) distances from $D^*$ to $P_c$ (in blue), from $D^*$ to $P^*$ (in red), and from $D^*$ to a uniformly random permutation in $\Pi$ (in black). Note that all three distances are very similar for $\rho<0.8$, implying that $D^*$ is very close to the barycenter and far from the boundary of $\mathcal{D}$.  With this in mind, it is not surprising that the projection fails to recover $P^*$ for $\rho<0.8$ in Figure~\ref{fig:energy}, as at the barycenter, the projection onto $\Pi$ is uniformly random.

For very high correlation values ($\rho>0.9$), the distances to $P_c$ and to $P^*$ sharply decrease, and the distance to a random permutation sharply increases.  This suggests that at these high correlation levels $D^*$ moves away from the barycenter and towards $P^*$.  Indeed, in Figure \ref{fig:energy} we see for $\rho>0.9$ that $P^*$ is the closest permutation to $D^*$, and is typically recovered by the projection step.



\subsection{On indefinite relaxed graph matching problem}
\label{sec:indef}

The continuous problem one would like to solve, $\min_{D \in {\mathcal D}} -\langle A'D,DB \rangle$ (since its optimum is $P^*$ with high probability), is indefinite. One option is to look for a local minimum of the objective function, as done in the FAQ algorithm of \cite{FAQ}.  The FAQ algorithm uses F-W methodology (\cite{FW}) to find a local minimum of $-\langle A'D,DB \rangle$.  Not surprisingly (as there are many local minima), the performance of the algorithm is heavily dependent on the initialization.  Below we study the effect of initializing the algorithm at the non-informative barycenter, at $D^*$ (a principled starting point), and at $P^*$.  We then compare performance of the different FAQ initializations to the PATH algorithm \cite{Zaslavskiy2009} and to the GLAG algorithm \cite{fiori2013nips}. 

The GLAG algorithm presents an alternate formulation of the graph matching problem.  The algorithm convexly relaxes the alternate formulation, solves the relaxation and projects it onto $\Pi$.  As demonstrated in \cite{fiori2013nips}, the algorithm's main advantage is in matching weighted graphs and multimodal graphs. The PATH algorithm begins by finding $D^*$, and then solves a sequence of concave and convex problems in order to improve the solution. The PATH algorithm can be viewed as an alternative way of projecting $D^*$ onto $\Pi$.  Together with FAQ, these algorithms achieve the current best performance in matching a large variety of graphs (see \cite{fiori2013nips}, \cite{FAQ}, \cite{Zaslavskiy2009}).  However, we note that GLAG and PATH often have significantly longer running times than FAQ (even if computing $D^*$ for FAQ:$D^*$); see \cite{FAQ,lyzinski2014spectral}.

\begin{figure}[t!]
\centering
\def\svgwidth{270pt}
\begin{scriptsize}
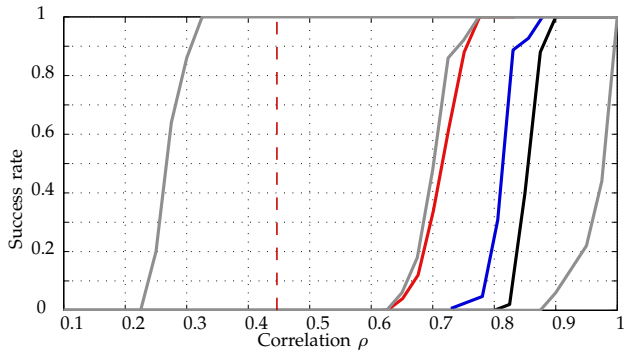
\end{scriptsize}
\caption{Success rate in recovering $P^*$. In gray, FAQ starting at, from left to right, $P^*$, $D^*$, and $J$; in black, $P_c$; in red, PATH; in blue, GLAG.  For each $\rho,$ we ran $100$ MC replicates.}
\label{fig:nonconvex}
\end{figure}

Figure \ref{fig:nonconvex} shows the success rate of the graph matching methodologies in recovering $P^*$. The vertical dashed red line at $\rho=0.44$ corresponds to the threshold in Theorem \ref{thd} part {\it a} (above which $P^*$ is optimal whp) for the parameters used in these experiments, and the solid lines correspond to the performance of the different methods: from left to right in gray, FAQ:$P^*$, FAQ:$D^*$, FAQ:$J$; in black, the success rate of $P_c$; the performance of GLAG and PATH are plotted in blue and red respectively.

Observe that, when initializing with $P^*$, the fact that FAQ succeeds in recovering $P^*$ means that $P^*$ is a local minimum, and the algorithm did not move from the initial point. From the theoretical results, this was expected for $\rho>0.44$, and the experimental results show that this is also often true for smaller values of $\rho$. However, this only means that $P^*$ is a local minimum, and the function could have a different global minimum. On the other hand, for very lowly correlated graphs ($\rho<0.3$), $P^*$ is not even a local minimum.

\begin{figure}[t!]
\def\svgwidth{270pt}
\begin{scriptsize}
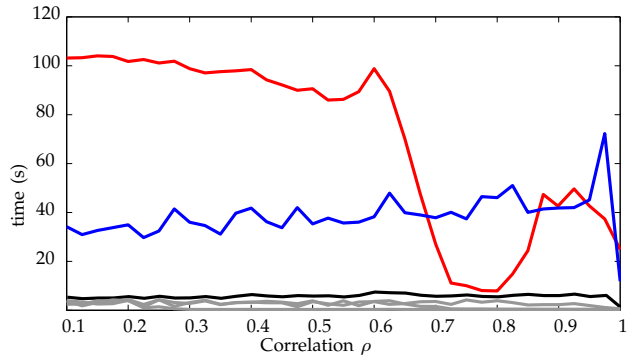
\end{scriptsize}
\caption{Average run time for FAQ:$D^*$ (note that this does not include the time to find $D^*$) and FAQ:$J$ in gray; finding $P_c$ (first finding $D^*$) in black; PATH in red; and GLAG in blue.  For each $\rho$, we average over 100 MC replicates.  Note that the runtime of PATH drop precipitously at $\rho=0.6,$ which corresponds to the performance increase in Figure \ref{fig:nonconvex}.}
\label{fig:times2}
\end{figure} 

The difference in the performance illustrated by the gray lines indicates that the resultant graph matching solution can be improved by using $D^*$ as an initialization to find a local minimum of the indefinite relaxed problem.  We see in the figure that FAQ:$D^*$ achieves best performance, 
while being computationally less intensive than PATH and GLAG, see Figure \ref{fig:times2} for the runtime result.
This amalgam of the convex and indefinite methodologies (initialize indefinite with the convex solution) is an important tool for obtaining solutions to graph matching problems, providing a computationally tractable algorithm with state-of-the-art performance.  

However, for all the algorithms there is still room for improvement.  In these experiments, for $\rho\in[0.44,0.7$ theory guarantees that with high probability the global minimum of the indefinite problem is $P^*$, and we cannot find it with the available methods. 

\begin{figure}[t!]
\hspace{-10mm}
\def\svgwidth{270pt}
\begin{scriptsize}
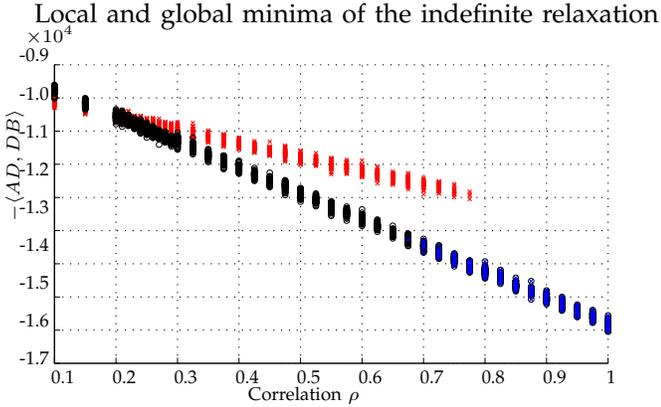
\end{scriptsize}
\caption{Value of $-\langle A'D,DB \rangle$ for $D=P^*$ (black) and for the output of FAQ:$D^*$ (red/blue indicating failure/success in recovering the true permutation).  For each $\rho,$ we ran $100$ MC replicates.}
\label{fig:trace}
\end{figure}

When FAQ:$D^*$ fails to recover $P^*$, how close is the objective function at the obtained local minima to the objective function at $P^*$?  Figure \ref{fig:trace} shows $-\langle A'D,DB \rangle$ for the true permutation, $P^*$, and for the pre-projection doubly stochastic local minimum found by FAQ:$D^*$. For $0.35<\rho<0.75$, the state-of-the-art algorithm not only fails to recover the correct bijection, but also the value of the objective function is relatively far from the optimal one. There is a transition (around $\rho\approx 0.75$) where the algorithm moves from getting a wrong local minimum to obtaining $P^*$ (without projection!).  For low values of $\rho$, the objective function values are very close, suggesting that both $P^*$ and the pre-projection FAQ solution are far from the true global minima.  At $\rho\approx0.3,$ we see a separation between the two objective function values (agreeing with the findings in Figure \ref{fig:nonconvex}).  As $\rho>0.44$, we expect that $P^*$ is the global minima and the pre-projection FAQ solution is far from $P^*$ until the phase transition at $\rho\approx0.75$.
\subsection{Real data experiments}
\label{data}
We further demonstrate the applicability of our theory in a series of real data examples.  First we match three pairs of graphs where a latent alignment is known.  We further compare different graph matching approaches on a set of 16 benchmark problems (those used in \cite{FAQ,Zaslavskiy2009}) from the QAPLIB QAP library of \cite{qaplib}, where no latent alignment is known a priori.  Across all of our examples, an intelligently initialized local solution of the indefinite relaxation achieves best performance.  

Our first example is from human connectomics.  For $45$ healthy patients, we have DT-MRI scans from one of two different medical centers: $21$ patients scanned (twice) at the Kennedy Krieger Institute (KKI), and
$24$ patients scanned (once) at the Nathan Kline Institute (NKI) (all data available at \url{ http://openconnecto.me/data/public/MR/MIGRAINE_v1_0/}).  Each scan is identically processed via the MIGRAINE pipeline of \cite{wrg1} yielding a $70$ vertex weighted symmetric graph.
In the graphs, vertices correspond to regions in the Desikan brain atlas, which provides the latent alignment of the vertices.
Edge weights count the number of neural fiber bundles connecting
the regions.  We first average the graphs within each medical center and then match the averaged graphs across centers.  

For our second example, the graphs consist of the two-hop neighborhoods of the ``Algebraic Geometry'' page in the French and English Wikipedia graphs.  The 1382 vertices correspond to Wikipedia pages with (undirected) edges representing hyperlinks between the pages.  Page subject provides the latent alignment function, and to make the graphs of commensurate size we match the intersection graphs. 

Lastly, we match the chemical and electrical connectomes of the C. elegans worm.  The connectomes consist of 253 vertices, each representing a specific neuron (the same neuron in each graph).  Weighted edges representing the strength of the (electrical or chemical) connection between neurons.  Additionally, the electrical graph is directed while the chemical graph is not.  
\begin{table}[t!]
\vspace{-4mm}
\caption{$\|A'P-PB\|_F$ for the $P$ given by each algorithm together with the number of vertices correctly matched ($n_{corr.}$) in real data experiments} 
\begin{tabular}{c | c | c c c} 
\hline 
{\bf Algorithm} & & {\bf KKI-NKI} & {\bf Wiki.} & {\bf C. elegans}\\ [0.5ex] 
\hline 
Truth & $\|A'P-PB\|_F$ &82892.87 & 189.35 & 155.00 \\
& $n_{corr.}$& 70 & 1381 & 253\\ \hline \hline
Convex relax. & $\|A'P-PB\|_F$&104941.16 & 225.27 & 153.38 \\
& $n_{corr.}$& 41 &97 &2\\ \hline
GLAG & $\|A'P-PB\|_F$& 104721.97 & 219.98 & 145.53 \\
& $n_{corr.}$& 36&181&4\\ \hline
PATH & $\|A'P-PB\|_F$&165626.63 & 252.55 & 158.60 \\
&$n_{corr.}$&1&1&1\\
\hline
FAQ:$J$ & $\|A'P-PB\|_F$ &93895.21 & 205.28 & 127.55\\
&$n_{corr.}$& 38&30&1\\ \hline
{\bf FAQ:D$^*$} & {\bf $\|A'P-PB\|_F$}& {\bf 83642.64} & {\bf 192.11} & {\bf 127.50} \\
& {\bf $n_{corr.}$}&{\bf 63}&{\bf 477}&{\bf 5}\\ 
\hline 
\end{tabular}
\label{table:nonlin} 
\end{table}

The results of these experiments are summarized in Table \ref{table:nonlin}.  In each example, the computationally inexpensive FAQ:$D^*$ procedure achieves the best performance compared to the more computationally expensive GLAG and PATH procedures.  This reinforces the theoretical and simulation results presented earlier, and again points to the practical utility of our amalgamated approach.  While there is a canonical alignment in each example, the results point to the potential use of our proposed procedure (FAQ:$D^*$) for measuring the strength~of this alignment, i.e., measuring the strength of the correlation between the graphs.  If the graphs are strongly aligned, as in the KKI-NKI example, the performance of FAQ:$D^*$ will be close to the truth and a large portion of the latent alignment with be recovered.  As the alignment is weaker, FAQ:$D^*$ will perform even better than the true alignment, and the true alignment will be poorly recovered, as we see in the C. elegans example. 

What implications do our results have in graph matching problems without a natural latent alignment?  To test this, we matched 16 particularly difficult examples from the QAPLIB library of \cite{qaplib}.  
We choose these particular examples, because they were previously used in \cite{FAQ,Zaslavskiy2009} to assess and demonstrate the effectiveness of their respective matching procedures.  
Results are summarized in Table \ref{fig:table}.  
We see that in every example, the indefinite relaxation (suitably initialized) obtains the best possible result.  Although there is no latent alignment here, if we view the best possible alignment as the ``true'' alignment here, then this is indeed suggested by our theory and simulations.  As the FAQ procedure is computationally fast (even initializing FAQ at {\it both} $J$ and $D^*$ is often comparatively faster than GLAG and PATH; see \cite{FAQ} and \cite{lyzinski2014spectral}), these results further point to the applicability of our theory.  Once again, theory suggests, and experiments confirm, that approximately solving the indefinite relaxation yields the best matching results.

\begin{table}[t!]
\vspace{-4mm}
\caption{$\|A'P-PB\|^2_F$ for the different tested algorithms on 16 benchmark examples of the QAPLIB library.}
\hspace{-2mm}
\includegraphics[width=0.50\textwidth]{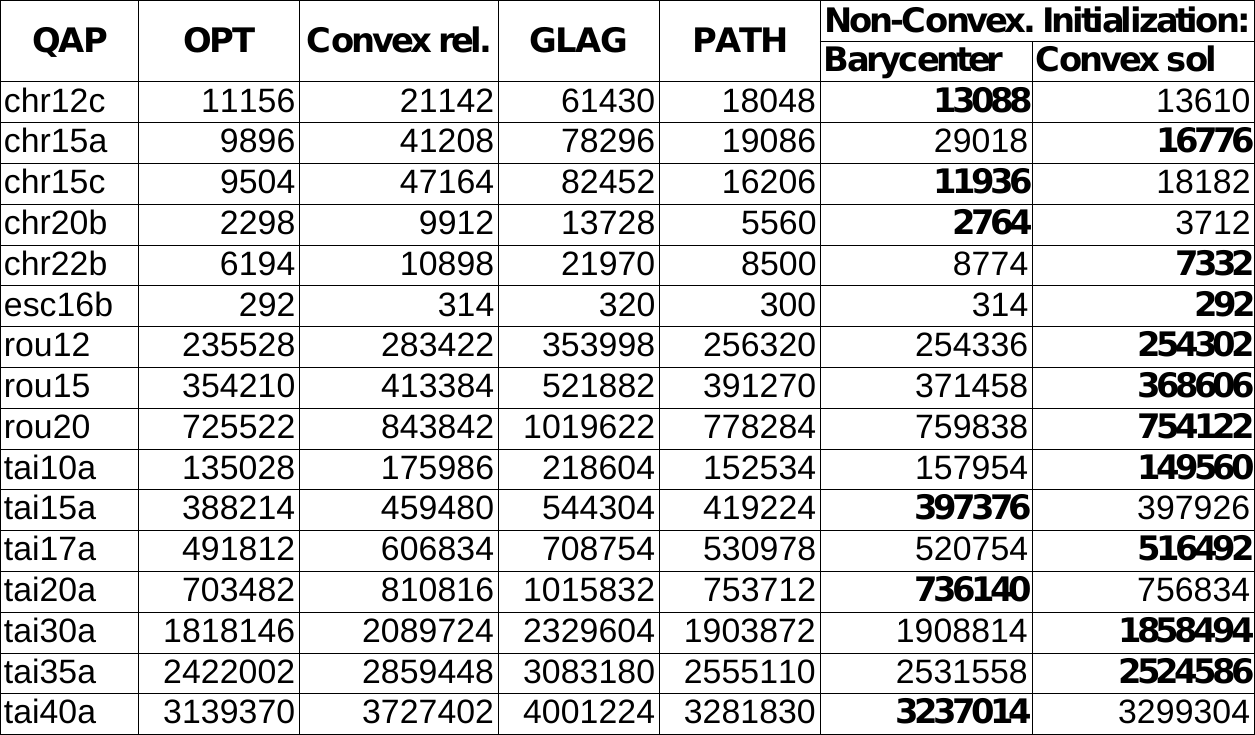}
\vspace{-5mm}
\label{fig:table}
\end{table} 

\subsection{Other random graph models}
\label{sec:otherrand}

While the random Bernoulli graph model is the most general edge-independent random graph model, in this section we present analogous experiments for a wider variety of edge-dependent random graph models. 
For these models, we are unaware of a simple way to exploit pairwise edge correlation in the generation of these graphs, as was present in Section \ref{sec:crbg}.  
Here, to simulate aligned non-isomorphic random graphs, we proceed as follows.  
We generate a graph $G_1$ from the appropriate underlying distribution, and then model $G_2$ as an errorful version of $G_1$; i.e., for each edge in $G_1$, we randomly flip the edge (i.e., bit-flip from $0\mapsto1$ or $1\mapsto0$) independently with probability $p\in[0,1]$.  
We then graph match $G_1$ and $G_2$, and we plot the performance of the algorithms in recovering the latent alignment function across a range of values of $p$.

We first evaluate the performance of our algorithms on \textit{power law} random graphs \cite{barabasi1999emergence}; these graphs have a degree distribution that follows a power law, i.e., the proportion of vertices of degree $d$ is proportional to $d^{-\beta}$ for some constant $\beta>0$.
These graphs have been used to model many real data networks, from the Internet \cite{albert1999internet,faloutsos1999power}, to social and biological networks \cite{girvan2002community}, to name a few.
In general, these graphs have only a few vertices with high degree, and the great majority of the vertices have relatively low degree.

\begin{figure}[t!]
\centering
\def\svgwidth{280pt}
\begin{scriptsize}
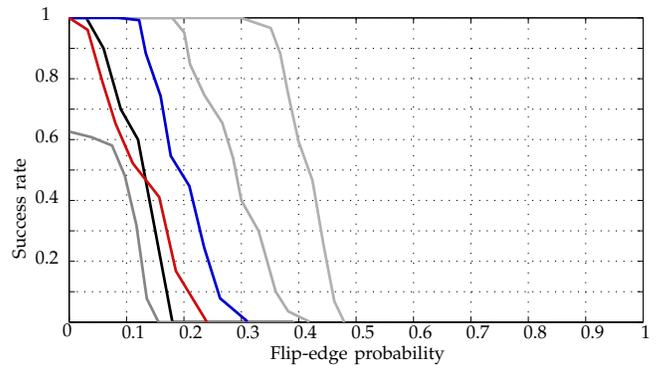
\end{scriptsize}
\caption{Success rate in recovering $P^*$ for $150$ vertex power law graphs with $\beta=2$ for: In gray, from right to left, FAQ:$P^*$, FAQ:$D^*$, and FAQ:$J$; in black, $P_c$; in red, PATH; in blue, GLAG.  For each value of the bit-flip parameter $p,$ we ran $100$ MC replicates.}
\label{fig:powerlaw}
\end{figure}

Figure \ref{fig:powerlaw} shows the performance comparison for the methods analyzed above: FAQ:$P^*$, FAQ:$D^*$, FAQ:$J$, $P_c$, PATH, and GLAG. For a range of $p\in[0,1]$, we generated a 150 vertex power law graph with $\beta=2$, and subsequently graph matched this graph and its errorful version.  For each $p$, we have 100  MC replicates. 
As with the random Bernoulli graphs, we see from Figure \ref{fig:powerlaw} that the true permutation is a local minimum of the non-convex formulation for a wide range of flipping probabilities ($p\leq0.3$), implying that in this range of $p$, $G_1$ and $G_2$ share significant common structure.
Across all values of $p<0.5$, FAQ:$P^*$ outperforms all other algorithms considered (with FAQ:$D^*$ being second best across this range).  This echoes the results of Sections (\ref{sec:conv})--(\ref{data}), and suggests an analogue of Theorem \ref{thd} may hold in the power law setting.  We are presently investigating this.

\begin{figure}[t!]
\vspace{-4mm}
\centering
Success rates for bounded degree graphs
\vspace{3mm}
\def\svgwidth{123pt}
\begin{tiny}
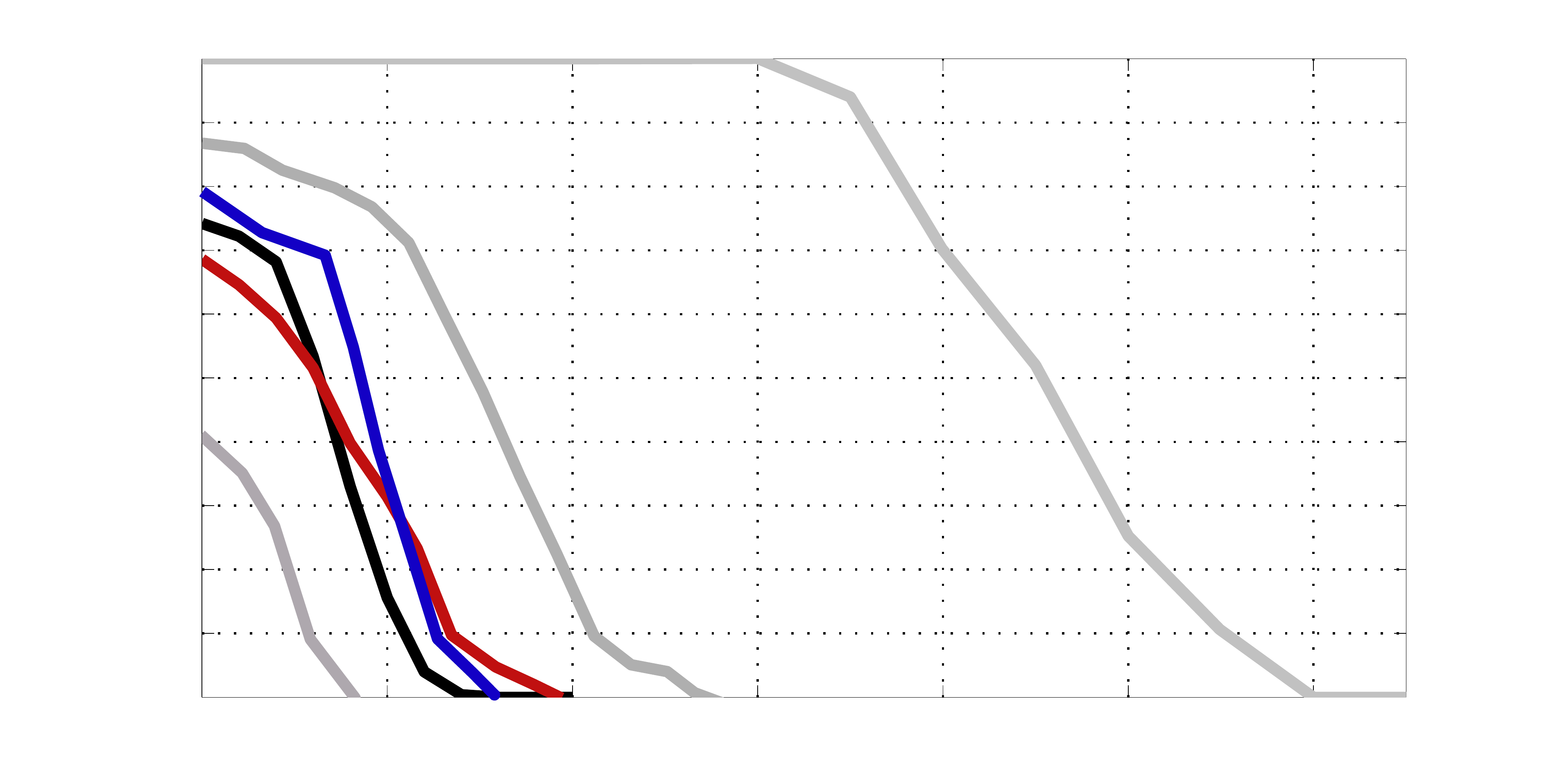
\def\svgwidth{123pt}
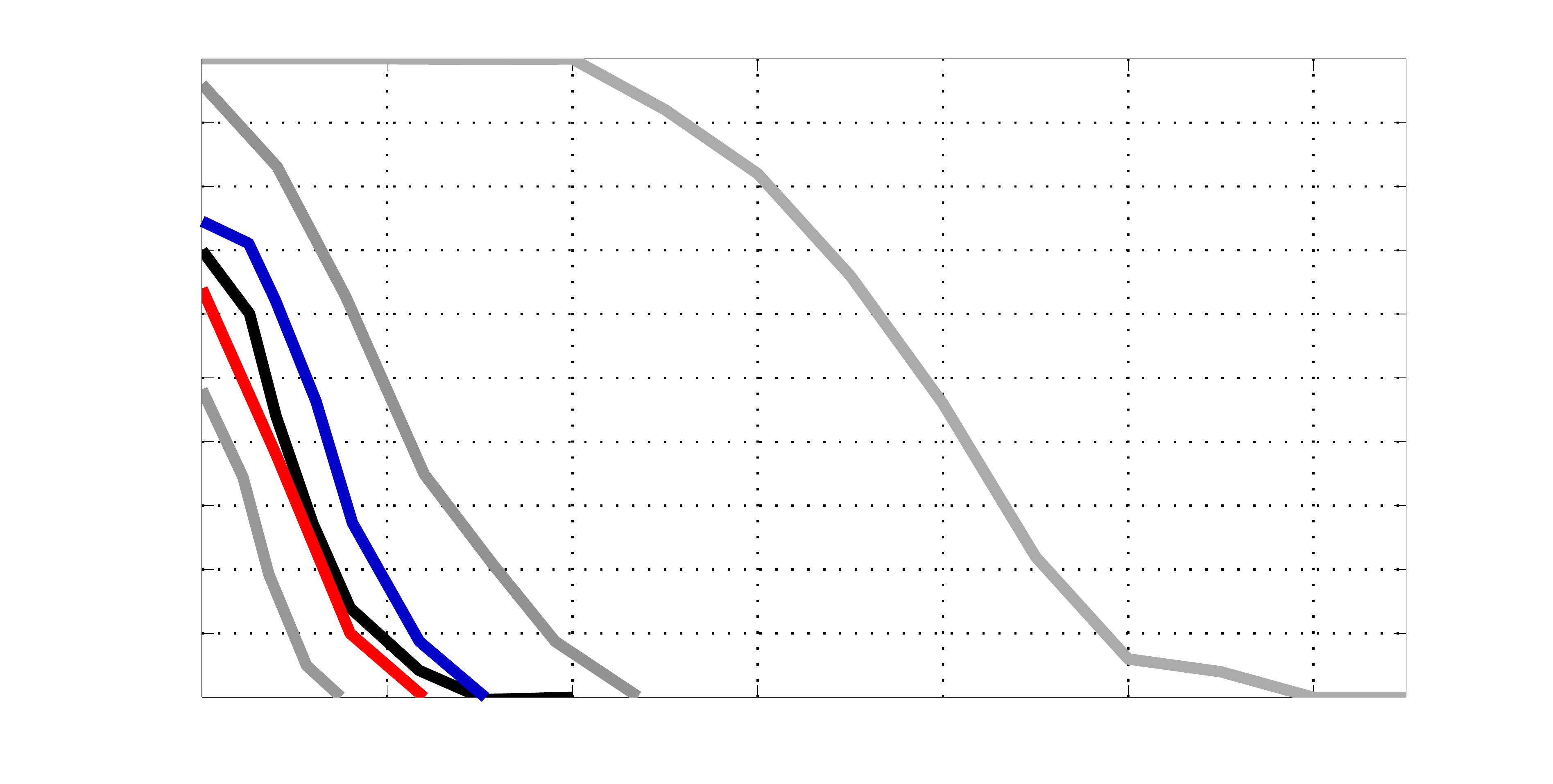
\vspace{0.4cm}
\def\svgwidth{123pt}
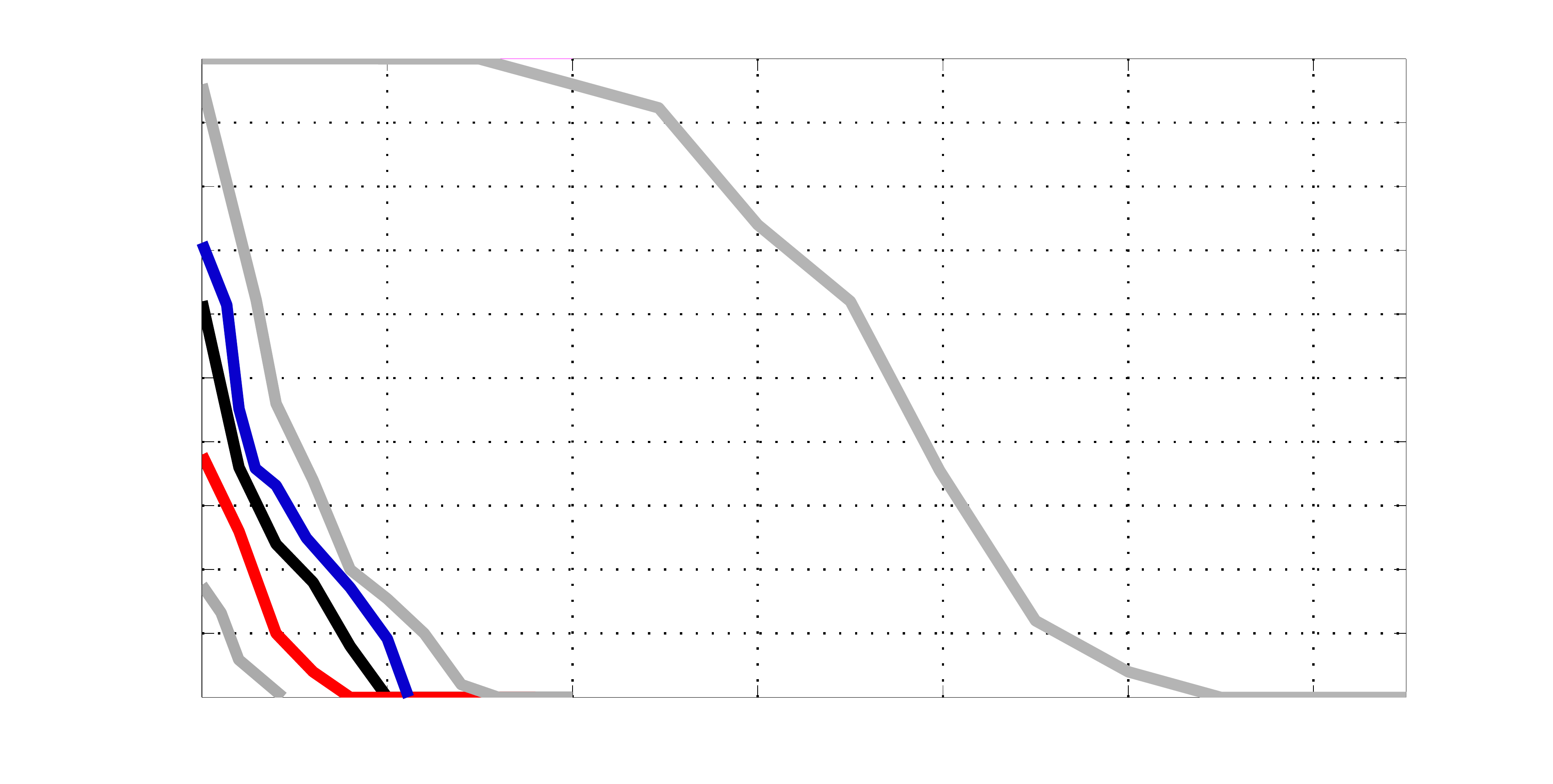
\def\svgwidth{123pt}
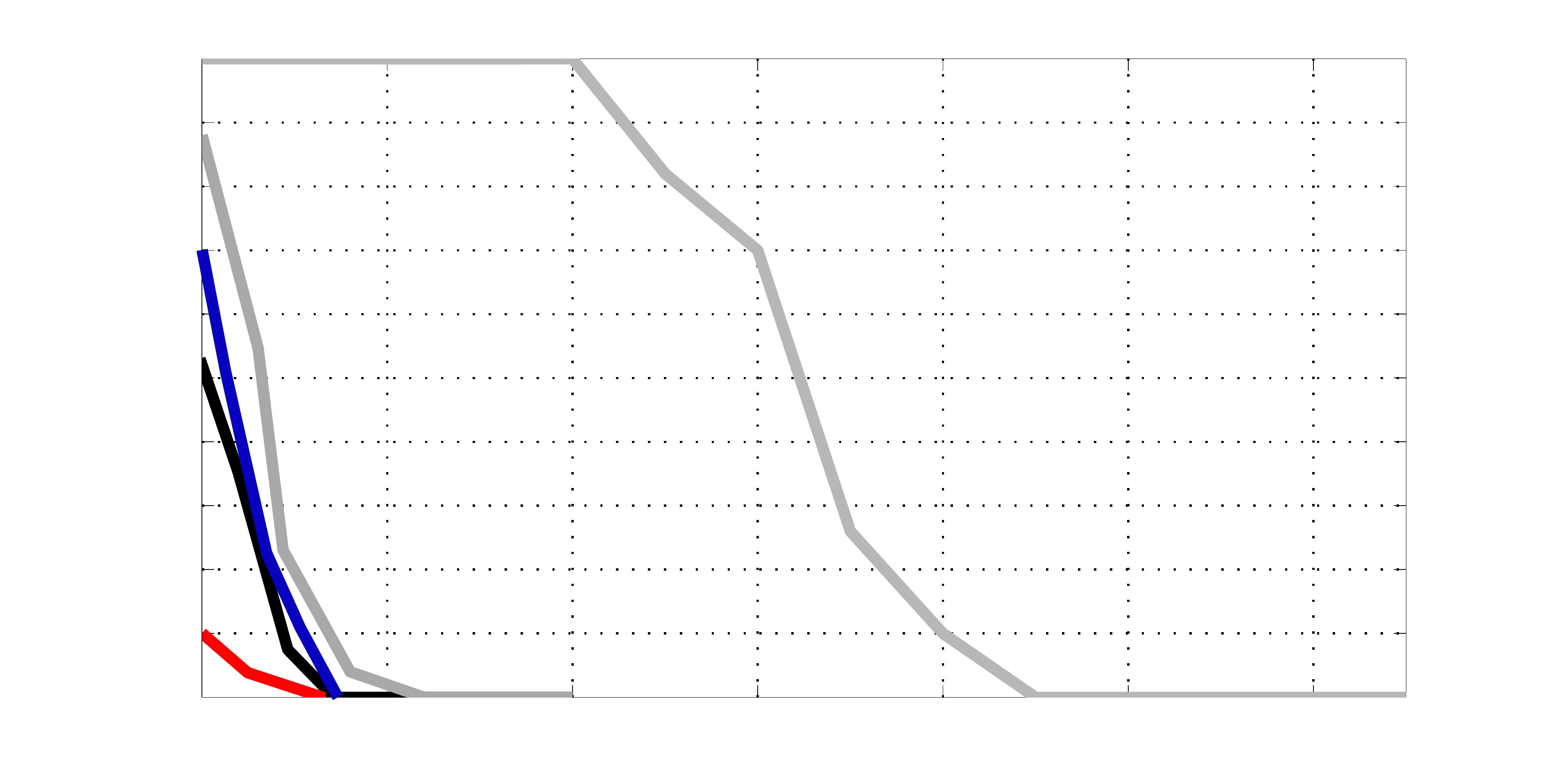
\end{tiny}
\caption{Success rate in recovering $P^*$ for bounded degree graphs (max degree $4$).  In gray, from right to left, FAQ:$P^*$, FAQ:$D^*$, and FAQ:$J$; in black, $P_c$;  in red, PATH; in blue, GLAG.  For each probability we ran $100$ MC replicates.}
\label{fig:bounded}
\end{figure}

We next evaluate the performance of our algorithms on graphs with bounded maximum degree (also called {\it bounded valence graphs}). These graphs have been extensively studied in the literature, and for bounded valence graphs, the graph isomorphism problem is in $P$ \cite{luks1982isomorphism}.
For the experiments in this paper we generate a random graph from the model in \cite{balinska2001algorithms} with maximum degree equal to $4$, and vary the graph order from $50$ to $350$ vertices. Figure \ref{fig:bounded} shows the comparison of the different techniques and initializations for these graphs, across a range of bit-flipping parameters $p\in[0,1]$.

It can be observed that even for isomorphic graphs ($p=0$), all but FAQ:$P^*$ fail to perfectly recover the true alignment. 
We did not see this phenomena in the other random graph models, and this can be explained as follows. It is a well known fact that convex relaxations fail for regular graphs \cite{fiori2014spectral}, and also that the bounded degree model tends to generate almost regular graphs \cite{koponen2012random}. Therefore, even without flipped edges, the graph matching problem with the original graphs is very ill-conditioned for relaxation techniques. Nevertheless, the true alignment is a local minimum of the non-convex formulation for a wide range of values of $p$ (shown by FAQ:$P^*$ performing perfectly over a range of $p$ in Figure \ref{fig:bounded}).  We again note that FAQ:$D^*$ outperforms $P_c$, PATH and GLAG across all graph sizes and bit-flip parameters $p$.  This suggests that a variant of Theorem \ref{thd} may also hold for bounded valence graphs as well, and we are presently exploring this.

We did not include experiments with any random
graph models that are highly regular and symmetric (for example, mesh graphs).  Symmetry and regularity have two effects on the graph matching problem. 
Firstly, it is well known that $P_c\neq P^*$ for non-isomorphic regular graphs (indeed, $J$ is a solution of the convex relaxed graph matching problem).
Secondly, the symmetry of these graphs means that there are potentially several isomorphisms
between a graph and its vertex permuted analogue.
Hence, any 
flipped edge could make permutations other than $P^*$ into the minima of the graph matching problem.

\subsection{Directed graphs}

\begin{figure}[t!]
\centering
\def\svgwidth{260pt}
\begin{scriptsize}
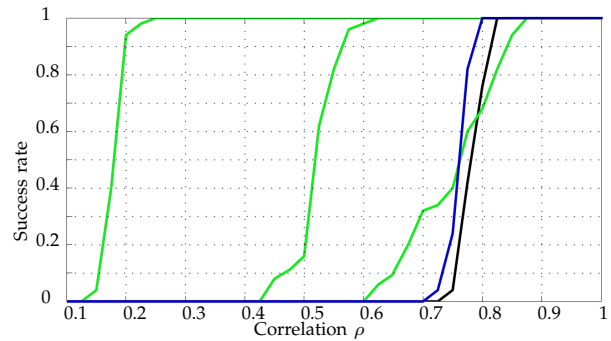
\end{scriptsize}
\caption{Success rate for directed graphs. We plot $P_c$ (black), the GLAG method (blue), and the nonconvex relaxation starting from different points in green, from right to left: FAQ:$J$, FAQ:$D^*$, FAQ:$P^*$.}
\label{fig:directed}
\end{figure}

All the theory developed above is proven in the undirected graph setting (i.e., $A$ and $B$ are assumed symmetric). However, directed graphs are common in numerous applications. Figure \ref{fig:directed} repeats the analysis of Figure \ref{fig:nonconvex} with directed graphs, all other simulation parameters being unchanged. The PATH algorithm is not shown in this new figure because it is designed for undirected graphs, and its 
performance for directed graphs is very poor. 
Recall that in Figure \ref{fig:nonconvex}, i.e., in the undirected setting, FAQ:$J$ performed significantly worse than $P_c$. 
In Figure \ref{fig:directed}, i.e., the directed setting, we note that the performance of FAQ:$J$ outperforms $P_c$ over a range of $\rho\in[0.4,0.7$].
As in the undirected case, we again see significant performance improvement (over FAQ:$J$, $P_c$, and GLAG) when starting FAQ from $D^*$ (the convex solution).  
Indeed, we suspect that a directed analogue of Theorem \ref{thd} holds, which would explain the performance increase achieved by the nonconvex relaxation over $P_c$.  
Here, we note that the setting for the remainder of the examples considered is the undirected graphs setting.

\subsection{Seeded graphs}

In some applications it is common to have some \textit{a priori} information about partial vertex correspondences, and seeded graph matching includes these known partial matchings as constraints in the optimization (see \cite{ModFAQ,JMLR:v15:lyzinski14a,alex}). However, seeds do more than just reducing the number of unknowns in the alignment of the vertices.
Even a few seeds can dramatically increase performance graph matching performance, and (in the $\rho$-correlated Erd\H os-R\'enyi setting) a logarithmic (in $n$) number of seeds contain enough signal in their seed--to--nonseed adjacency structure to a.s. perfectly align two graphs \cite{JMLR:v15:lyzinski14a}. 
Also, as shown in the deterministic graph setting in \cite{alex}, very often $D^*$ is closer to $P^*$.  

\begin{figure}[t!]
\centering
\def\svgwidth{270pt}
\begin{scriptsize}
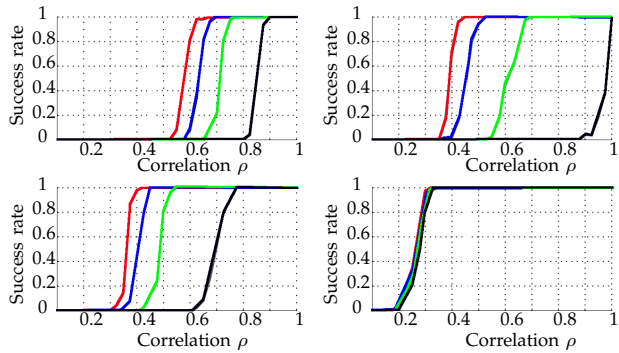
\end{scriptsize}
\caption{Success rate of different methods using seeds. We plot $P_c$ (top left), FAQ:$J$ (top right), FAQ:$D^*$ (bottom left), and FAQ:$P^*$ (bottom right). For each method, the number of seeds increases from right to left: 0 (black), $5$ (green), $10$ (blue) and $15$ (red) seeds.  Note that more seeds increases the success rate across the board.}
\label{fig:seeds}
\end{figure}

In Figure \ref{fig:seeds}, the graphs are generated from the $\rho$-correlated random Bernoulli model with random $\Lambda$ (entrywise uniform over $[0.1,0.9]$). We run the Frank-Wolfe method (modified to incorporate the seeds) to solve the convex relaxed graph matching problem, and the method in \cite{ModFAQ,JMLR:v15:lyzinski14a} to approximately solve the nonconvex relaxation, starting from $J$, $D^*$, and $P^*$.  Note that with seeds, perfect matching is achieved even below the theoretical bound on $\rho$ provided in Theorem 1 (for ensuring $P^*$ is the global minimizer). 
This provides a potential way to improve the theoretical bound on $\rho$ in Theorem \ref{thd}, and the extension of Theorem 1 for graphs with seeds is the subject of future research.
With the exception of the nonconvex relaxation starting from $P^*$, each of the different 
FAQ initializations and the convex formulation all see significantly improved performance as the number of seeds increases.  We also observe that the nonconvex relaxation seems to benefit much more from seeds than the convex relaxation.
Indeed, when comparing the performance with no seeds, the $P_c$ performs better than FAQ:$J$. 
However, with just five seeds, this behavior is inverted.  
Also of note, in cases when seeding returns the correct permutation, we've empirically observed that merely initializing the FAQ algorithm with the seeded start, and not enforcing the seeding constraint, also yields the correct permutation as its solution (not shown).

\begin{figure}[t!]
\centering
\def\svgwidth{280pt}
\begin{scriptsize}
\input{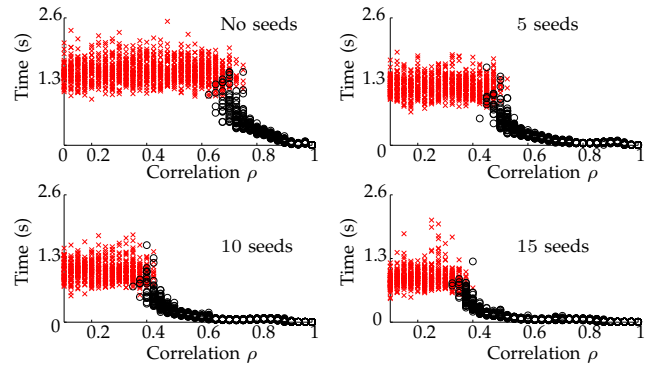}
\end{scriptsize}
\caption{Running time for the nonconvex relaxation when starting from $D^*$, for different number of seeds. A red ``x'' indicates the algorithm failed to recover $P^*$, and a black ``o'' indicates it succeeded. In each, the algorithm was run to termination at discovery of a local min.}
\label{fig:times}
\end{figure}

Figure \ref{fig:times} shows the running time (to obtain a solution) when starting from $D^*$ for the nonconvex relaxation, using different numbers of seeds. For a fixed seed level, the running time is remarkably stable across $\rho$ when FAQ does not recover the true permutation.  On the other hand, when FAQ does recover the correct permutation, the algorithm runs significantly faster than when it fails to recover the truth.  This suggests that, across all seed levels, the running time might, by itself, be a good indicator of whether the algorithm succeeded in recovering the underlying correspondence or not.  Also note that as seeds increase, the overall speed of convergence of the algorithm decreases and, unsurprisingly, the correct permutation is obtained for lower correlation levels.

\subsection{Features}

Features are additional information that can be utilized to improve performance in graph matching methods, and often these features are manifested as additional vertex characteristics besides the connections with other vertices.
For instance, in social networks we may have have a complete profile of a person in addition to his/her social connections.

We demonstrate the utility of using features with the nonconvex relaxation, the standard convex relaxation and the GLAG method, duely modified to include the features into the optimization. Namely, the new objective function to minimize is
$
\lambda F(P) + (1-\lambda)\textup{trace}(C^TP), \, 
$ 
where $F(P)$ is the original cost function ($-\langle AP,PB\rangle$ in the nonconvex setting, $\|AP-PB\|_F^2$ for the convex relaxation and $\sum_{i,j}\|( [AP]_{i,j},[PB]_{i,j})\|_2$ for the GLAG method), the matrix $C$ codes the features fitness cost, and the parameter $\lambda$ balances the trade-off between pure graph matching and fit in the features domain. For each of the matching methodologies, the optimization is very similar to the original featureless version.

For the experiments, we generate $\rho$-correlated Bernoulli graphs as before, and in addition we generate a Gaussian random vector (zero mean, unit variance) of $5$ features for each node of one graph, forming a $5\times n$ matrix of features; we permute that matrix according to $P^*$ to align new features vectors with the nodes of the second graph. Lastly, additive zero-mean Gaussian noise with a range of variance values is added to each feature matrix independently.  If for each vertex $v\in[n]$ the resulting noisy feature for $G_i$, $i=1,2$, is $x_v^{(i)}$, then the entries of $C$ are defined to be 
$C_{v,w}=\|x_v^{(1)}-x_w^{(2)}\|_2,$
for $v,w\in[n]$.
Lastly, we set $\lambda=0.5.$

\begin{figure}[t!]
\centering
\def\svgwidth{260pt}
\begin{scriptsize}
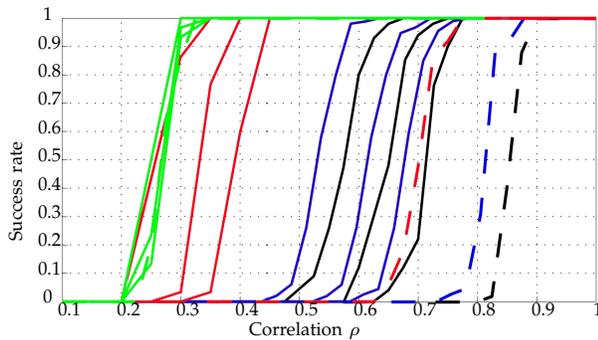
\end{scriptsize}
\caption{Success rate of different methods using features: $P_c$ (in black), GLAG (in blue), FAQ:$D^*$ (in red), and FAQ:$P^*$ (in green). For each method, the noise level (variance of the Gaussian random noise) increases from left to right: $0.3$, $0.5$, and $0.7$. In dashed lines, we show the success of the same methods without features.}
\label{fig:features}
\end{figure}

Figure \ref{fig:features} shows the behavior of the methods when using features for different levels of noise in the feature matrix. 
Even for highly noisy features (recalling that both feature matrices are contaminated with noise), this external information still helps in the graph matching problem. 
For all noise levels, all three methods improve their performance with the addition of features, and of course, the improvement is greater when the noise level decreases.  
Note that, as before, $FAQ$ outperforms both $P_c$ and GLAG across all noise levels.
It is also worth noting that for low noise, FAQ:$D^*$ performs comparably to FAQ:$P^*$, which we did not observe in the seeded (or unseeded) setting.

Even for modestly errorful features, including these features improves downstream matching performance versus the setting without features.
This points to the utility of high fidelity features in the matching task. 
Indeed, given that the state-of-the-art graph matching algorithms may not achieve the optimal matching for even modestly correlated graphs, the use of external information like seeds and features can be critical.

\section{Conclusions}
In this work we presented theoretical results showing the surprising fact that the indefinite relaxation (if solved exactly) obtains the optimal solution to the graph matching problem with high probability,  under mild conditions.  Conversely, we also present the novel result that the popular convex relaxation of graph matching almost always fails to find the correct (and optimal) permutation.  In spite of the apparently negative statements presented here, these results have an immediate practical implication: the utility of intelligently initializing the indefinite matching algorithm to obtain a good approximate solution of the indefinite problem.   

The experimental results further emphasize the trade-off between tractability and correctness in relaxing the graph matching problem,  
with real data experiments and simulations in non edge-independent random graph models suggesting that our theory could be extended to more general random graph settings.  Indeed, 
all of our experiments corroborate that best results are obtained via approximately solving the intractable indefinite problem.  
Additionally, both theory and examples point to the utility of combining the convex and indefinite approaches, using the convex to initialize the indefinite.  


\bibliographystyle{IEEEtran}
\bibliography{biblio}
\end{document}